%% file: main.tex
\documentclass{article}

\usepackage[OT1]{fontenc}
\usepackage{iclr2027_conference,times}
\input{math_commands.tex}

\usepackage{amsmath}
\usepackage{amssymb}
\usepackage{booktabs}
\usepackage{graphicx}
\usepackage{microtype}
\usepackage{url}
\usepackage{placeins}
\usepackage{float}
\usepackage{wrapfig}
\usepackage{hyperref}
\hypersetup{hidelinks,bookmarksopen=true,bookmarksnumbered=true,
    pdftitle={A Spectral Theory of Compositional Learning},pdfauthor={Hugo Rydel}}
\newcommand{\appref}[1]{\hyperref[#1]{Appendix~\ref*{#1}}}
\newcommand{\paperfigref}[1]{\hyperref[#1]{Figure~\ref*{#1}}}
\newcommand{\figpanelref}[2]{\hyperref[#1]{Figure~\ref*{#1}#2}}
\newcommand{\papertabref}[1]{\hyperref[#1]{Table~\ref*{#1}}}

\newtheorem{proposition}{Proposition}[section]

\title{A Spectral Theory of Compositional Learning}

\author{Hugo Rydel\\\normalfont University of Manchester}

\usepackage{etoolbox}
\makeatletter
\patchcmd{\@maketitle}
  {\begin{tabular}[t]{l}\bf\rule{\z@}{24pt}\@author\end{tabular}}
  {\hbox to \textwidth{\hfil\begin{tabular}[t]{c}\bf\rule{\z@}{24pt}\@author\end{tabular}\hfil}}
  {}{\PackageError{author-version}{Could not center the author block}{}}
\makeatother

\iclrfinalcopy

\begin{document}
\raggedbottom

\maketitle
\lhead{In review at ICLR 2027}

\begin{abstract}
How does compositional reasoning emerge during learning? We address this question by mathematically analyzing the learning dynamics of deep linear networks. We train these networks in structured synthetic environments and derive a theory linking the structure of experience to compositional learning. Our theory predicts when compositional inferences emerge, whether they are identifiable from the available evidence, and how new linking evidence can rapidly unlock previously unavailable inferences. These results provide a qualitative explanation for several phenomena observed in human cognition. They account for why a composition can fail despite knowing its premises, why similar compositions can emerge at different times, and how a single linking fact can suddenly enable many new inferences. Taken together, these findings establish a mathematical link between the statistical structure of experience and the development of compositional reasoning.
\end{abstract}

\section{Introduction}
\label{sec:introduction}

Compositional reasoning is fundamental to human intelligence. It allows us to combine known relations into new ones, extending our knowledge beyond direct experience \citep{frankland2020two,halford2010relational}. For example, if we know that $a>b$ and $b>c$, we can infer that $a>c$ without observing that fact. This capacity supports reasoning across domains, from interpreting social and spatial relationships to understanding language and solving unfamiliar problems \citep{behrens2018cognitive}.

Compositional reasoning, however, does not arise uniformly across contexts. In associative-inference tasks, individuals can learn overlapping relational premises such as A--B and B--C yet still remain poor at grasping the unseen A--C composition \citep{zeithamova2010flexible, kumaran2012generalization}. How those relations are experienced also matters. When learning cleanly isolates the structural components of premises, compositional generalization improves and can emerge through intermediate partial solutions \citep{dekker2022curriculum}. By contrast, when two relational structures have already been learned separately, only a small amount of linking evidence can rapidly unlock many compositional inferences \citep{nelli2023neural}. Thus, similar premise knowledge can result in persistent compositional failure, gradual generalization, or sudden inference depending on how relational evidence is organized and presented. So, what determines which of these trajectories occurs?

Mathematical theories of learning offer a way to explain these differences. In deep linear networks, the training data can be decomposed into separate components, and each component is learned at a speed set by how strongly the data express it \citep{saxe2014exact,saxe2019mathematical,lampinen2019analytic}. These components and their strengths make up the spectrum of the training data, which determines which parts of a representation are learned and when. These analyses concern outputs that a network is trained to produce, whereas a compositional inference is, by definition, never trained. However, a composition combines relations that are each learned from training facts, so it can only become available once those relations are learned. Its timing should therefore depend on how strongly the training facts constrain those relations, just as the strength of a component sets how quickly it is learned. If the training facts leave one of these relations undetermined, further training on the same facts should not be able to settle it, and the composition remains unavailable until a new fact is added. When several compositions depend on the same undetermined relation, a single new fact that determines it can make all of them available at once. If this rule holds, gradual generalization, persistent compositional failure, and sudden inference could all arise from the same learning dynamics.

We test this account in synthetic relational worlds with known additive laws, using shallow and deep linear neural networks. Our contributions are as follows:
\begin{itemize}
    \item In Section~\ref{sec:identifiability}, we give a criterion for whether the observed facts uniquely determine a compositional law, even when the full embedding remains undetermined (Proposition~\ref{prop:identifiability}).
    \item In Sections~\ref{sec:emergence}--\ref{sec:depth}, we derive closed-form shallow learning trajectories as sums of decaying spectral modes (Proposition~\ref{prop:dynamics}) and numerically integrate the coupled dynamics of deeper networks to predict compositional retrieval.
    \item In Section~\ref{sec:results}, we test these predictions in three experiments and three network depths. Across independently generated worlds, the predictions track the trained networks' retrieval and geometry, showing how evidence governs law-specific emergence and how linking facts can resolve previously unavailable inferences.
\end{itemize}

\section{Related Work}
\label{sec:related}

\paragraph{Relational representations.}
Relational embeddings can encode compositional rules by making a sequence of relations act like a single relation. Translation models give this an additive form by representing relations as vector displacements \citep{bordes2013translating}. More broadly, training can produce algebraic relationships between relation representations that reflect compositional patterns in the data \citep{guu2015traversing,sun2019rotate}. These learned relationships are the object of our analysis. In additive worlds, we characterize which observed facts require composed displacements to agree with a single relation and predict how that agreement develops during learning.

\paragraph{Structural identifiability.}
Training examples can admit several interpretations that disagree on unseen combinations. Identification-based theories establish conditions under which training coverage and architectural constraints make compositional structure recoverable without observing every combination \citep{wiedemer2023compositional,schug2024discovering}. Thus, they show that, for relational inference, different embeddings can satisfy the same observations and enforce the same compositional law while disagreeing in others. We build on this knowledge in Proposition~\ref{prop:identifiability}, where we characterize when a particular law should hold across all embeddings consistent with the evidence.

\paragraph{Compositional learning dynamics.}
The structure of training evidence influences both the order and course of generalization. Spectral analyses connect learning times to the recovery of structured representations \citep{saxe2019mathematical,liu2022understanding}, while interactions between learning directions can produce temporary reversals in compositional performance \citep{yang2025swing}. Our analysis contributes to this literature by deriving the mean learning dynamics from the relational training objective and using it to follow individual laws through learning. Reconstructing the evolving dynamics yields predictions for their retrieval trajectories, including after linking evidence changes the constraints.

\paragraph{Hidden progress and sudden emergence.}
Successful generalization can become visible only after substantial internal progress. Mechanistic analyses of grokking and studies of concept learning reveal developing computational structure and capabilities that ordinary performance measures initially conceal \citep{nanda2023progress,park2024emergence}. Their findings motivate distinguishing progress toward an inference from successful retrieval. Our geometric error measures how closely a query approaches its target, including while retrieval remains incorrect. Predicting this error alongside accuracy lets us examine how changes in the learned geometry precede and produce observable compositional success.

\paragraph{Knowledge integration.}
Learning a new fact need not make its consequences available. Knowledge-editing benchmarks demonstrate that language models can reproduce an edited fact while failing to update answers that depend on it \citep{zhong2023mquake,cohen2024evaluating}. Human knowledge-assembly experiments show how consequential such propagation can be, with limited linking information enabling many inferences across separately learned structures \citep{nelli2023neural}. Our linking experiment isolates this problem in a consistent additive system. We identify the shared ambiguity removed by a linking fact and predict how its consequences become retrievable, including temporary disruption of previously learned premises.

\section{Experimental Framework}
\label{sec:framework}

\subsection{Relational worlds and learning model}

\begin{wrapfigure}{r}{0.40\linewidth}
    \centering
    \vspace{-0.9\baselineskip}
    \includegraphics[width=\linewidth]{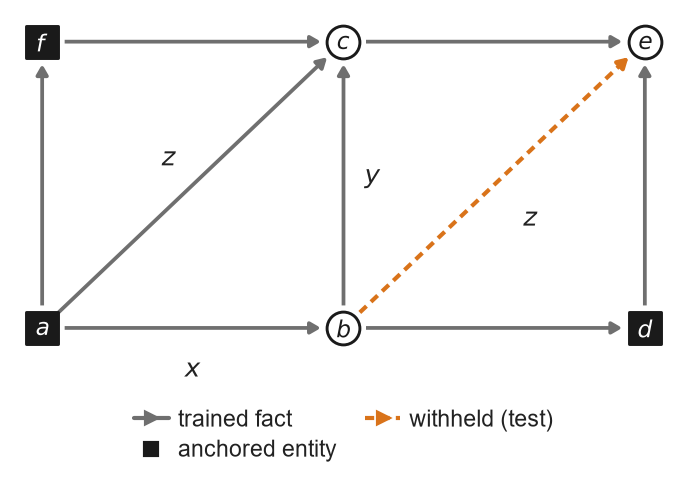}
    \caption{Example relational world.}
    \label{fig:world}
\end{wrapfigure}

Our worlds consist of entities connected by directed relations (Fig.~\ref{fig:world}). A fact $a\xrightarrow{r}b$ states that entity $a$ stands in relation $r$ to entity $b$. All labels are arbitrary, so any structure must be learned from the facts themselves.

\setlength{\columnsep}{16pt}  

The model places each entity $a$ at a learned position $\mathbf{x}_a\in\mathbb{R}^d$ and represents each relation $r$ by a learned displacement $\mathbf{r}\in\mathbb{R}^d$, shared by every fact that uses $r$. Learning $a\xrightarrow{r}b$ means making this displacement carry $a$ to $b$, so training minimizes the squared norm of each fact's error,
\begin{equation}
    \boldsymbol{\epsilon}_{a,r,b}=\mathbf{x}_a+\mathbf{r}-\mathbf{x}_b.
    \label{eq:fact}
\end{equation}

Composition is then addition of displacements. If the model has learned $a\xrightarrow{x}b$ and $b\xrightarrow{y}c$, following both relations from $a$ gives $\mathbf{x}_a+\mathbf{r}_x+\mathbf{r}_y\approx\mathbf{x}_c$. Our worlds contain a third relation $z$ that joins the two ends of every such path, so one $z$ step must cover the same displacement as the two-step path: $\mathbf{r}_z=\mathbf{r}_x+\mathbf{r}_y$. This compositional law holds in the world by construction but is not built into the model, which must recover it from the facts. Each law spans many entities; training presents only some of the facts that hold, and each experiment withholds others as tests. Because facts constrain only differences between positions, the whole configuration could otherwise shift freely. Anchor penalties fix it by drawing a few entities toward fixed, distinct coordinates, while leaving them learnable.

We collect all positions and displacements as rows of an embedding matrix $E\in\mathbb{R}^{p\times d}$, one row per entity or relation token. The shallow model ($N=1$) learns $E$ directly, as a lookup table. A depth-$N$ model instead factorizes it as $E=W_N\cdots W_1$, with one $p\times d$ factor and $N-1$ factors of size $d\times d$. This changes its learning dynamics without changing representational capacity. We use $d=16$ and $N=1,2,3$, and train by stochastic gradient descent (SGD) on individual facts and anchors, reshuffled each epoch. \appref{app:training} gives initialization, learning rates, and training durations.

\subsection{Experimental manipulations}

\begin{figure}[H]
    \centering
    \includegraphics[width=\linewidth]{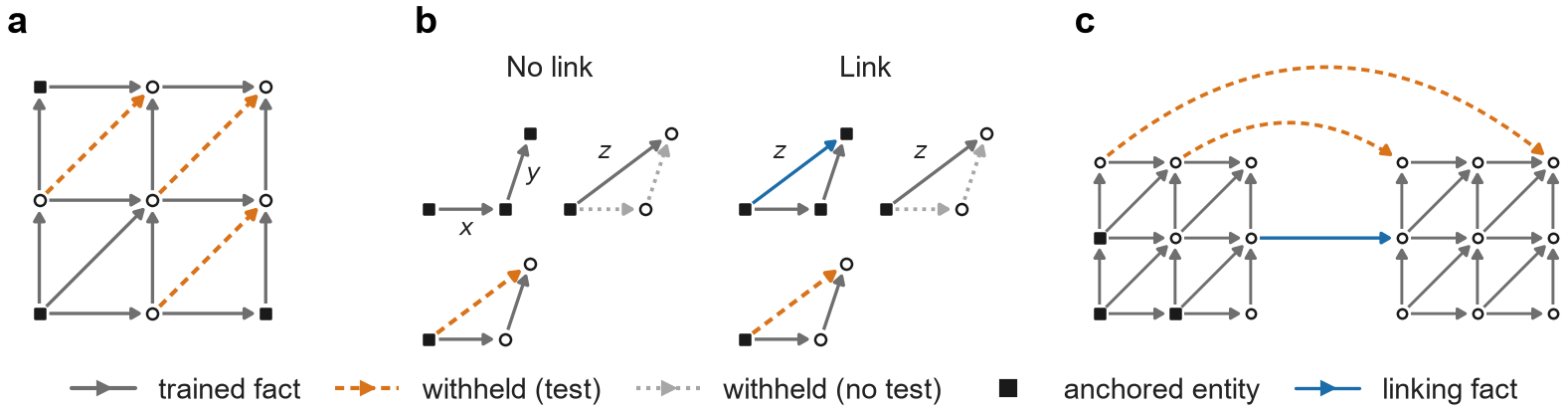}
    \caption{Experimental manipulations of compositional learning.}
    \label{fig:experiments}
\end{figure}

\paragraph{Experiment 1: law-specific emergence.}
Networks learn 16 compositional laws supported by lattices of different sizes and fact-presentation frequencies (Fig.~\ref{fig:experiments}a). Each lattice contains observed and withheld $z$-facts. We test whether differences in the supporting evidence predict when each law's withheld compositions become retrievable.

\paragraph{Experiment 2: structural identifiability.}
Observing all three relations need not determine their composition. When observed $z$-facts connect entity pairs separate from the observed $x$-then-$y$ paths, the facts leave the law $\mathbf{r}_z=\mathbf{r}_x+\mathbf{r}_y$ undetermined (Fig.~\ref{fig:experiments}b). After pretraining, we add one $z$-fact connecting the endpoints of an observed path and test whether other withheld compositions become retrievable. A comparison structure contains such closing facts from the outset.

\paragraph{Experiment 3: relational integration.}
Networks first learn two lattices with different entities but shared relation vectors (Fig.~\ref{fig:experiments}c). Only the first lattice is anchored, leaving their relative position undetermined. We then connect them with one fact using an existing relation and test inference between other cross-lattice entity pairs.

In Experiments~2 and 3, the \emph{linking fact} is one additional observed relation, trained with the same error objective as the original facts. Linked and no-link conditions begin from identical pretrained weights. The linked condition revisits the added fact each epoch alongside the original facts; the no-link condition continues with the original facts alone. \appref{app:worlds} specifies the constructions, and \appref{app:training} gives the intervention schedules.

\subsection{Measuring compositional learning}
\label{sec:measures}

We evaluate inference on facts and comparisons withheld from training. In Experiments~1 and 2, a query $a\xrightarrow{z}b$ places the model's prediction at $\mathbf{q}=\mathbf{x}_a+\mathbf{r}_z$. In Experiment~3, the query applies supplied numbers of $x$ and $y$ steps from a source entity to predict a target in the other lattice (see \appref{app:evaluation}).

\emph{Compositional accuracy} measures how often the correct target is the nearest candidate entity to the query in Euclidean distance. Candidates belong to the relevant lattice or block in Experiments~1 and 2, and to both lattices in Experiment~3. \emph{Geometric error} measures the mean query-to-target distance relative to the current entity spacing in the target's structure. This captures continuous changes that may precede a change in retrieval accuracy. \appref{app:evaluation} gives the query sets and precise metric definitions.

Theoretical predictions start from the same initial weights as the networks and use the same evaluation queries, without fitting their observed trajectories. We summarize results across 200 worlds per depth, treating the world as the unit of replication. \appref{app:statistics} defines emergence criteria and summary statistics. Separate retention tests track retrieval of previously trained facts after each intervention (\appref{app:controls}).

\section{A Spectral Theory of Compositional Learning}
\label{sec:theory}

We ask which compositional laws the training evidence determines and how their retrieval develops during learning. The analysis gives a closed-form trajectory for shallow networks and coupled dynamics, integrated numerically, for deeper networks. \appref{app:derivations} provides the derivations.

\subsection{Representing the evidence}

Each training fact imposes a linear constraint on the embedding matrix $E$. For a fact $a\xrightarrow{r}b$, we construct a row of a matrix $A$ with coefficients $+1$ on the source entity and relation, $-1$ on the target entity, and zero elsewhere. Multiplying this row by $E$ gives the fact's error $\boldsymbol{\epsilon}_{a,r,b}$ (Equation~\ref{eq:fact}). An anchor row has a single $+1$ on the anchored entity, with its fixed coordinates stored in a target matrix $C$. Stacking the fact and anchor constraints gives the error matrix $AE-C$ and objective
\begin{equation}
    \mathcal{L}(E)=\frac{1}{2|D|}\lVert AE-C\rVert_F^2,
    \label{eq:loss}
\end{equation}
where $D$ is the multiset of training constraints. Repeated rows encode presentation frequency, with linking facts adding one new row. \appref{app:system} gives the full construction.

Two matrices determine the loss gradient, $\nabla_E\mathcal{L}=HE-B$:
\begin{equation}
    H=\frac{A^\top A}{|D|},
    \qquad
    B=\frac{A^\top C}{|D|}.
    \label{eq:hb}
\end{equation}
Here, $H$ captures the structure and weighting of the constraints, while $B$ incorporates the anchor targets. Since $H$ is symmetric and positive semidefinite, it admits the eigendecomposition $H = Q \Lambda Q^\top$, where the columns $q_k$ of $Q$ are orthonormal eigenvectors and $\Lambda$ contains their nonnegative eigenvalues $\lambda_k$.

Each \emph{spectral mode} $q_k\in\mathbb{R}^p$ specifies a combination of entity and relation embeddings, $q_k^\top E$. Its eigenvalue measures how strongly the training evidence constrains that combination. Positive eigenvalues identify constrained modes while zero eigenvalues identify changes to the embeddings that leave every training error unchanged. This distinction separates what the evidence determines from what remains free.

\subsection{Which laws does the evidence determine?}
\label{sec:identifiability}

For a law $j$, such as $\mathbf{r}_z=\mathbf{r}_x+\mathbf{r}_y$, define the contrast $\ell_j\in\mathbb{R}^p$ with coefficients $+1$ on relations $x$ and $y$, $-1$ on $z$, and zero elsewhere. The quantity $\ell_j^\top E=\mathbf{r}_x+\mathbf{r}_y-\mathbf{r}_z$ then measures the law's error. We call the law identifiable if this error is zero in every embedding matrix satisfying the training constraint $AE=C$.

Let $Q_+$ contain the eigenvectors of $H$ with positive eigenvalues. These modes span the subspace constrained by the evidence, and $Q_+Q_+^\top$ is the projection onto that subspace.

\begin{proposition}
\label{prop:identifiability}
If at least one solution of $AE=C$ satisfies the law, the law is identifiable if and only if
\begin{equation}
    \ell_j=Q_+Q_+^\top\ell_j.
    \label{eq:identifiability}
\end{equation}
\end{proposition}

This applies the classical estimability criterion for linear contrasts \citep{searle1971linear}. The contrast must lie entirely within the constrained subspace; otherwise, embeddings can fit every training fact while disagreeing about the composition. We quantify this ambiguity in Experiment~2 by
\begin{equation}
    \rho_j=\frac{\lVert\ell_j-Q_+Q_+^\top\ell_j\rVert_2}{\lVert\ell_j\rVert_2},
    \label{eq:rho-main}
\end{equation}
which is zero exactly when the law is identifiable.

A linking fact adds a row to $A$ that constrains a previously free direction. In both experiments it removes the only such direction. In Experiment~2, this makes the law identifiable. In Experiment~3, it fixes the relative position of two lattices, determining many cross-lattice comparisons simultaneously. \appref{app:identifiability} proves Proposition~\ref{prop:identifiability}, and \appref{app:linking} shows how a linking fact removes the unconstrained translation between the two lattices.

\subsection{When do identifiable laws emerge?}
\label{sec:emergence}

Even an identifiable law takes time to learn. For the shallow model, small-step SGD is approximated by the gradient flow
\begin{equation}
    \tau\frac{dE}{dt}=B-HE, \qquad \tau=(\eta\vert D\vert)^{-1},
    \label{eq:shallow}
\end{equation}
where $\eta$ is the per-constraint learning rate and $t$ is measured in epochs. Let $E^\ast$ satisfy $AE^\ast=C$, and define the embedding error along mode $k$ as
\begin{equation}
    \boldsymbol\zeta_k(t)=q_k^\top\bigl(E(t)-E^\ast\bigr).
    \label{eq:mode-error}
\end{equation}
Under this flow, constrained modes decay independently at rates $\lambda_k/\tau$, while unconstrained modes remain fixed \citep{saxe2014exact,saxe2019mathematical}. A law's error combines these modes through its loadings $a_{jk}=\ell_j^\top q_k$. By Proposition~\ref{prop:identifiability}, an identifiable law depends only on constrained modes, whose errors decay during learning.

\begin{proposition}
\label{prop:dynamics}
For an identifiable law $j$, its error under the shallow gradient flow is
\begin{equation}
    \boldsymbol\delta_j(t)=\ell_j^\top E(t)=\sum_{k:\lambda_k>0}a_{jk}\boldsymbol\zeta_k(0)e^{-\lambda_k t/\tau}.
    \label{eq:multimode}
\end{equation}
\end{proposition}

\begin{figure}[H]
    \centering
    \includegraphics[width=0.95\linewidth]{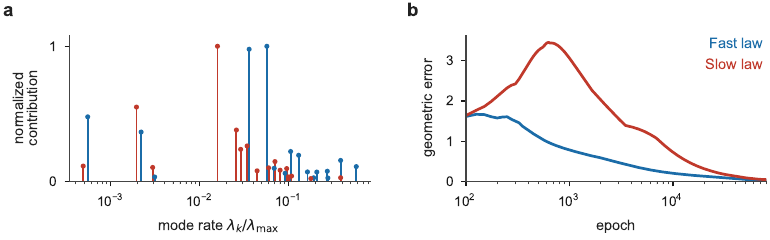}
    \caption{Fast and slow laws differ in spectral support. (a) Contribution of each spectral mode to the initial error of the fast and slow laws, normalized within each law and plotted against its relative decay rate, $\lambda_k/\lambda_{\max}$. (b) Predicted geometric error for the same laws in one $N=1$ world. The fast law places more weight on faster modes and therefore converges earlier than the slow law.}
    \label{fig:spectral}
\end{figure}

Laws supported by different modes can therefore emerge at different times within one network, with their trajectories dependent on the decay rates and the initial errors (Fig.~\ref{fig:spectral}).

Retrieval also depends on the positions of targets and competing candidates. We therefore reconstruct the full embedding trajectory and apply the measures from Section~\ref{sec:measures}, using a discrete mean-update predictor described in \appref{app:discrete}. \appref{app:shallow} provides the derivation of Proposition~\ref{prop:dynamics}.

\subsection{How does depth change learning?}
\label{sec:depth}

The factorization $E=W_N\cdots W_1$ changes how the embeddings are learned without restricting which embeddings the model can represent. Proposition~\ref{prop:identifiability} therefore applies at every depth.

To describe the factorized dynamics, define the products surrounding each factor $W_l$ as
\begin{equation}
    L_l=W_N\cdots W_{l+1}, \qquad R_l=W_{l-1}\cdots W_1,
\end{equation}
with empty products equal to the identity. Applying the chain rule to $\nabla_E\mathcal L=HE-B$ gives the factorized gradient flow \citep{saxe2014exact,lampinen2019analytic},
\begin{equation}
    \tau\frac{dW_l}{dt}=L_l^\top(B-HE)R_l^\top.
    \label{eq:deep}
\end{equation}
The evolving factors generally couple the spectral modes, so their contributions to a law's error can no longer be predicted independently. We integrate these equations numerically from the same initial weights as the trained network, reconstruct $E(t)$, and apply the retrieval measures from Section~\ref{sec:measures}.

Factorization also allows unconstrained modes to change during learning, whereas shallow dynamics preserve their initial values. Thus, depth can affect an underdetermined law through the solution selected by learning, even though the evidence still does not uniquely determine that law. \appref{app:depth} derives the factorized dynamics and explains when unconstrained modes can change.

\section{Results}
\label{sec:results}

\subsection{Spectral structure predicts law-specific emergence}
\label{sec:result1}

\begin{figure}[t]
    \centering
    \includegraphics[width=0.95\linewidth]{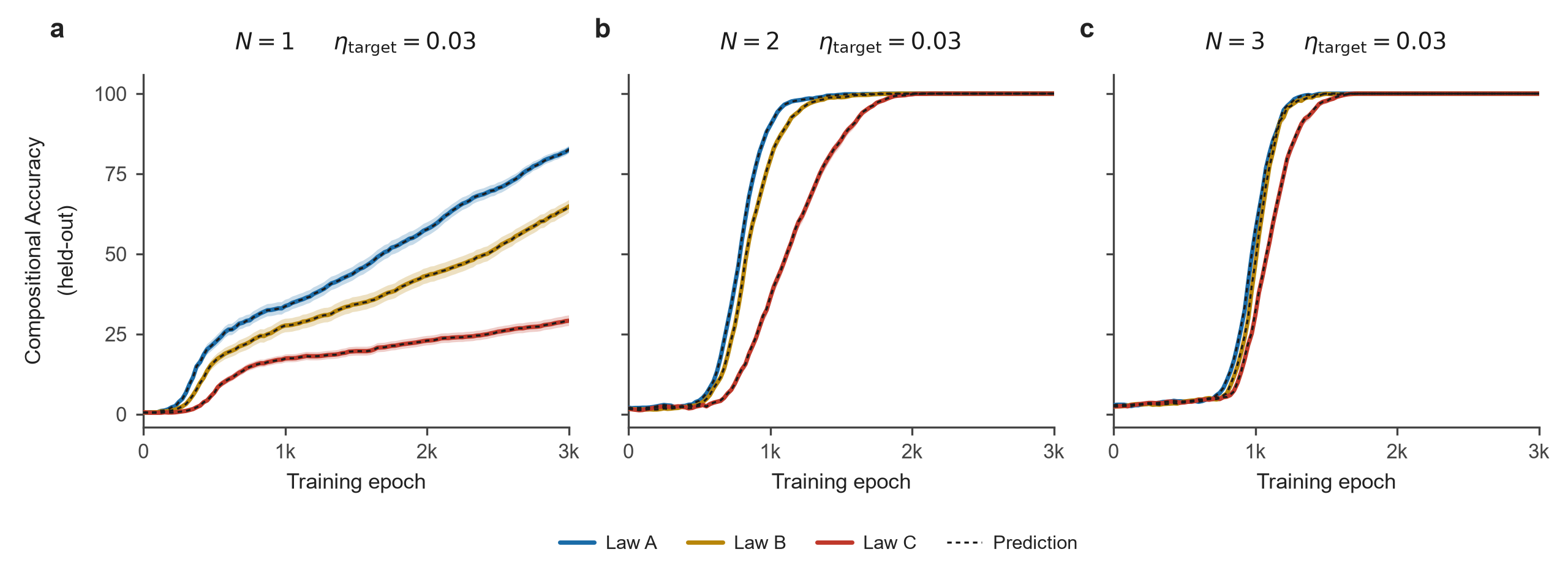}
    \caption{Compositional laws emerge at law-specific times predicted by theory. Three laws are shown. \textbf{a--c}, instantaneous accuracy on held-out queries $a\xrightarrow{z}b$. Shading is s.e.m. across 200 worlds. The first 3,000 epochs are shown; \appref{app:timing} gives full-run event times. The corresponding geometric trajectories are shown in \figpanelref{fig:geometry}{a--c}.}
    \label{fig:emergence}
\end{figure}

Compositional laws supported by different training evidence became retrievable on different timescales. Emergence time $t^\ast$ was the first evaluation at which held-out compositional accuracy reached 100\% and geometric error fell below 0.5, with both sustained for ten consecutive evaluations. Across the 16 laws, median $t^\ast$ over worlds spanned 4,000--21,438 epochs at $N=1$, 1,025--2,062 at $N=2$, and 1,125--1,400 at $N=3$, a 5.4-, 2.0-, and 1.2-fold range. The corresponding behavioural trajectories are shown in \paperfigref{fig:emergence}.

Starting from each network's initial weights, the theory predicted these differences in timing without fitting the observed trajectories. All 3,200 law--world pairs reached the emergence criterion in both the networks and their predictions at each depth, within 80,000 epochs at $N=1$ and 10,000 at $N=2,3$. Median absolute timing error was zero epochs at every depth, and 100.0\%, 99.9\%, and 98.8\% of predictions fell within one 25-epoch evaluation interval. Predicted and observed $\log_{10}$ emergence times agreed with $R^2$ = 1.0000, 0.9979, and 0.9793. Agreement extended to the full accuracy and geometric-error trajectories (\papertabref{tab:trajectories}).

\subsection{Resolving ambiguity enables compositional generalization}
\label{sec:result2}

\begin{figure}[H]
    \centering
    \includegraphics[width=0.95\linewidth]{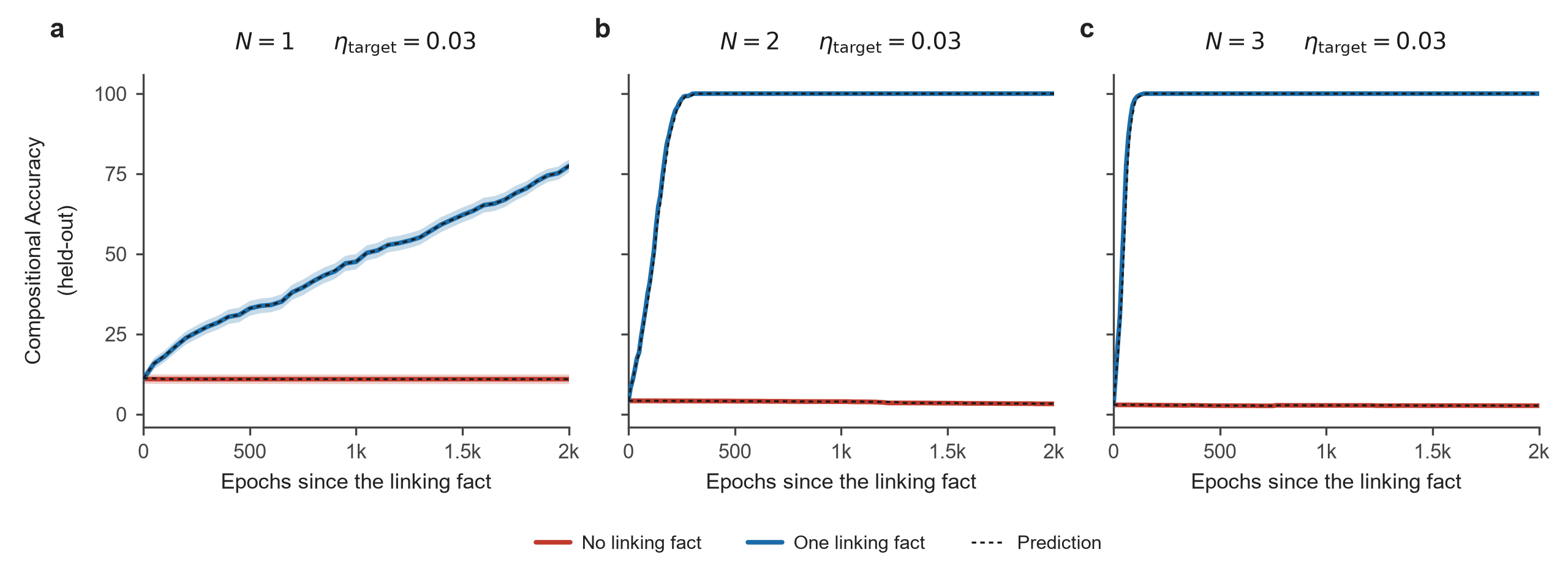}
    \caption{A single linking fact makes an underdetermined composition learnable. \textbf{a--c}, held-out accuracy. Shading is s.e.m. across 200 worlds. The corresponding geometric trajectories are shown in \figpanelref{fig:geometry}{d--f}.}
    \label{fig:identifiability}
\end{figure}

When the existing evidence left a law ambiguous, adding a single linking fact enabled generalization to withheld compositions. Before the intervention, the law's ambiguity was $\rho=0.2582$ in every world and depth; the linking fact reduced it to at most $1.4\times10^{-15}$. Two thousand epochs later, linked networks retrieved 77.4\%, 100.0\%, and 100.0\% of held-out compositions at $N=1,2,3$, compared with 10.9\%, 3.2\%, and 2.6\% in matched no-link controls continuing from the same pretrained weights (\paperfigref{fig:identifiability}). Every linked world reached 90\% accuracy, with median recovery times of 1,850, 160, and 60 epochs. Linked geometric error fell 1.82-, 29.7-, and 36.3-fold over the same period, whereas without the link accuracy changed by less than one percentage point and geometric error by at most 4\% (\figpanelref{fig:geometry}{d--f}).

The predicted dynamics captured how quickly compositional accuracy improved after the ambiguity was resolved. The median absolute error in recovery time was 0, 10, and 10 epochs, no more than one evaluation interval (50, 10, and 10 epochs), and 100.0\%, 100.0\%, and 99.5\% of predictions fell within one interval. The $R^2$ of $\log_{10}$ recovery time was 0.9986, 0.9860, and 0.8735. The lower $R^2$ at $N=3$ reflects reduced relative timing precision despite a median absolute error of only 10 epochs; \appref{app:agreement} examines this difference and reports agreement over the full accuracy and geometric-error trajectories.

To test whether the intervention disrupted earlier knowledge, we replayed 30 worlds at each depth under both structural assignments and both intervention conditions. All 72 original training facts, including the 64 premises, remained correct at every sampled evaluation over the full post-intervention horizons (\appref{app:controls}). Thus, the new compositions became retrievable without any detected loss of earlier knowledge on these evaluation grids.

\subsection{One linking fact unlocks cross-structure inference}
\label{sec:result3}

\begin{figure}[H]
    \centering
    \includegraphics[width=0.95\linewidth]{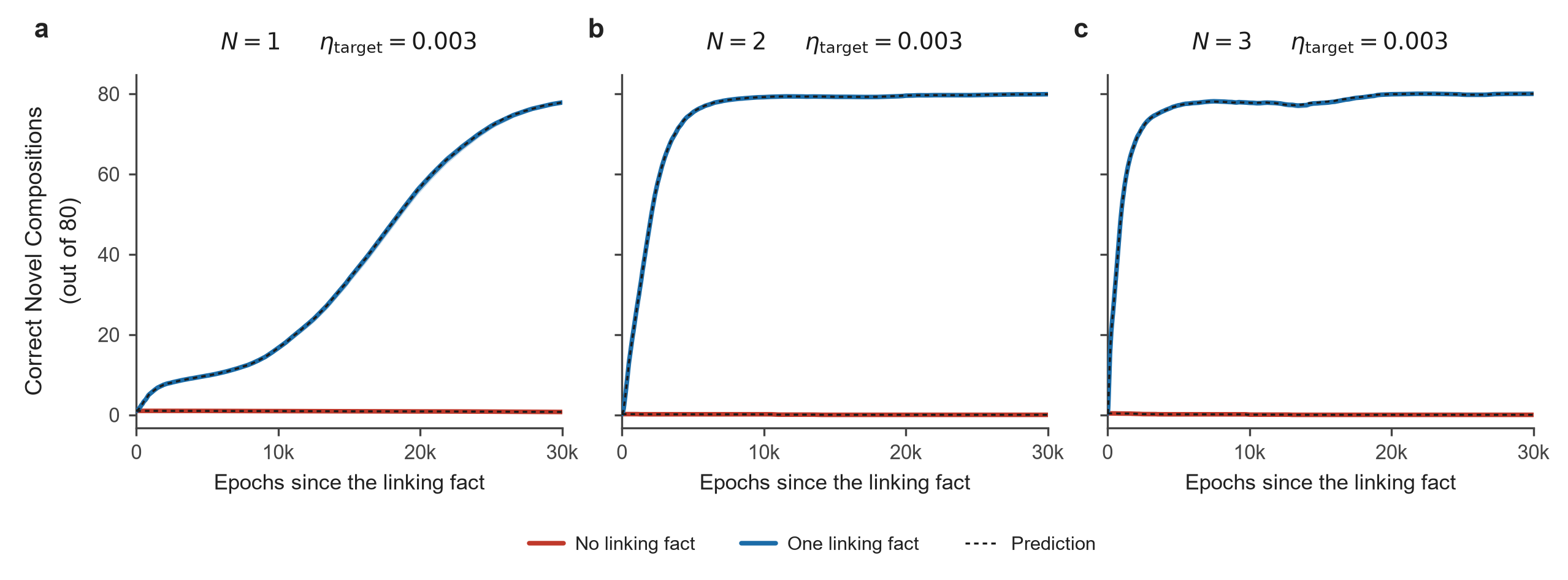}
    \caption{One linking fact unlocks many cross-structure inferences. \textbf{a--c}, absolute number of correctly answered held-out cross-structure comparisons, averaged across worlds. Shading is s.e.m. across 200 worlds. The corresponding geometric trajectories are shown in \figpanelref{fig:geometry}{g--i}.}
    \label{fig:integration}
\end{figure}

A linking fact also enabled generalization between two structures that had been learned separately. Before the link, networks answered almost none of the 80 held-out cross-structure comparisons: 1.0, 0.2, and 0.4 at $N=1,2,3$. Thirty thousand epochs later, linked networks answered 77.8, 79.9, and 80.0, compared with 0.7, 0.0, and 0.0 in matched no-link controls continuing from the same pretrained weights (\paperfigref{fig:integration}). Median time to 72 correct comparisons was 24,082, 3,771, and 2,250 epochs. Cross-structure geometric error fell 7.73-, 21.0-, and 163-fold with the link and changed by at most 5\% without it (\figpanelref{fig:geometry}{g--i}).

The same dynamics predicted how long these cross-structure inferences took to become retrievable. The median absolute error in the time to reach 72 correct comparisons was zero epochs at every depth, and 99.5\%, 100.0\%, and 95.0\% of predictions fell within the observed crossing interval, the gap between the last evaluation below criterion and the first at or above it (median 943, 155, and 58.5 epochs; \appref{app:statistics}). The $R^2$ of $\log_{10}$ unlock time was 0.9996, 0.9994, and 0.9996. The predictions also tracked the accuracy and geometric-error trajectories (\papertabref{tab:trajectories}).

This integration came with a largely temporary loss of previously learned premises (\paperfigref{fig:retention}). At least one of the 24 original premise facts became incorrect in 162, 170, and 136 of 200 linked worlds at depths 1--3; none of the no-link worlds showed such a loss. Of these 468 affected world--depth cases, 467 regained full premise retrieval at subsequent checks through the end of the run. One depth-1 world had not recovered by 40,000 epochs. The predicted and observed mean retention curves differed by at most 0.17 percentage points on the shared evaluation grids. Thus, the theory captured both the gain in cross-structure inference and the accompanying disruption of earlier knowledge.

\FloatBarrier

\section{Discussion and Limitations}
\label{sec:discussion}

We asked how training evidence determined which compositional laws a neural network could recover and when they became retrievable. In controlled additive worlds, our theory connected the constraints imposed by observed facts to the learning dynamics of relational embeddings. For identifiable laws, the theory predicted both their retrieval trajectories and emergence times across shallow and deep networks. When a law remained underdetermined, the theory identified the source of the ambiguity and predicted how retrieval developed after a linking fact resolved it. Linking two previously separate relational structures likewise enabled many cross-structure inferences, whose development was predicted by the same dynamics. Together, these findings show how the organization of training evidence can account for differences in compositional learning, from laws acquired at different times to coordinated gains in generalization after new evidence is introduced.

Our results distinguished two reasons why compositional retrieval could remain poor despite continued training. An identifiable law could remain poorly retrieved because the modes supporting it were still being learned. An underdetermined law instead remained compatible with conflicting solutions, even when every training constraint was satisfied. This distinction matters when interpreting held-out performance. A split may test whether a learner recovers a consequence of the observations, or whether its inductive bias selects an answer that the observations leave unresolved. In our setting, the identifiability criterion established which problem the learner faced before its trajectory was examined. It therefore provided a way to separate insufficient evidence from slow learning.

Our theory was developed in synthetic worlds with known additive laws and predefined relation types. This allowed us to isolate compositional learning, but left open whether the same principles apply to heterogeneous or non-additive relations. The analysis was also limited to shallow and deep linear networks. Although factorization introduced nonlinear learning dynamics, the experiments did not establish whether our predictions extend to nonlinear architectures. Future work should test whether the spectral structure of relational evidence remains informative in these settings, including on knowledge-graph benchmarks such as FB15k-237 \citep{toutanova2015observed} and WN18RR \citep{dettmers2018convolutional}. Matched relational tasks with human learners or nonlinear networks (e.g.\ large language models) could also provide a controlled test by comparing the empirical order of compositional emergence with our predictions. Such tests would clarify which aspects of our account generalize beyond the additive models studied here.

\clearpage
\subsection*{AI use statement}

Generative AI tools were used to help develop and refine the theoretical framing, formulate mathematical claims, design experiments, implement and debug analysis code, interpret results, search the literature, structure the manuscript, and edit prose. The author inspected the mathematical derivations, tested the code and numerical predictions against saved outputs, checked cited work, and reviewed all AI-assisted text. The author takes responsibility for the final content of this work, including claims, code, text, and artifacts produced with the aid of generative AI.

\subsection*{Ethics statement}

This study uses synthetic relational worlds and computational models. It involves no human participants, personal data, or animal subjects. We identify no direct ethical risks beyond the general obligation to report theoretical assumptions and computational results accurately.

\subsection*{Reproducibility statement}

The relational objective, theoretical assumptions, prediction equations, evaluation measures, and experimental manipulations are specified in Sections~\ref{sec:framework}--\ref{sec:theory}. Further derivations and procedural details appear in Appendices~\ref{app:derivations}--\ref{app:methods}.

\bibliography{references}
\bibliographystyle{iclr2027_conference}

\appendix
\numberwithin{figure}{section}
\numberwithin{table}{section}
\renewcommand{\theHfigure}{appendix.\thesection.\arabic{figure}}
\renewcommand{\theHtable}{appendix.\thesection.\arabic{table}}

\clearpage
\section{Mathematical Derivations}
\label{app:derivations}

We give the training-system construction, proofs of Propositions~\ref{prop:identifiability} and \ref{prop:dynamics}, and the extensions needed for finite learning rates, depth, and linking interventions. The algebraic identifiability result is exact. The trajectory results are exact for the stated mean dynamics, which approximate the per-row SGD used in the experiments.

\subsection{Constructing the training system}
\label{app:system}

Let $e_u\in\mathbb{R}^p$ select token $u$. Each row of $E$ stores the embedding of one entity or relation. A fact $a\xrightarrow{r}b$ contributes the row $(e_a+e_r-e_b)^\top$ to $A$ and a zero row to $C$. An anchor for entity $a$ with target $c_a\in\mathbb{R}^d$ and weight $w_a$ contributes $\sqrt{w_a}e_a^\top$ to $A$ and $\sqrt{w_a}c_a^\top$ to $C$. Thus Equation~\ref{eq:loss} is the sum of squared fact errors and weighted squared anchor errors, normalized by twice the number of rows. We use $w_a=1$. Anchors penalize deviations from a target; they do not clamp an embedding. Repeated facts contribute repeated rows, preserving their presentation frequencies.

The matrices have dimensions $A\in\mathbb{R}^{|D|\times p}$, $C\in\mathbb{R}^{|D|\times d}$, $H\in\mathbb{R}^{p\times p}$, and $B\in\mathbb{R}^{p\times d}$. Since $H=A^\top A/|D|$ is positive semidefinite, its positive-eigenvalue subspace is the row space of $A$. Write the orthogonal projectors onto this subspace and its complement as
\begin{equation}
    P_+=Q_+Q_+^\top,\qquad P_0=I-P_+.
    \label{eq:projectors}
\end{equation}
The complement is $\ker A$, the directions unobserved by training. Our generated systems are consistent: their constructed coordinates provide a solution to $AE=C$.

\subsection{Identifiability: proof of Proposition~\ref{prop:identifiability}}
\label{app:identifiability}

Fix a consistent solution $E^\star$ that satisfies the law, $\ell_j^\top E^\star=0$. Every other solution is $E=E^\star+V$ with $AV=0$. If $P_0\ell_j=0$, then $\ell_j^\top V=0$ for every such $V$, so all solutions satisfy the law.

Conversely, suppose $v=P_0\ell_j\ne0$. For any nonzero $u\in\mathbb{R}^d$, set $V=vu^\top$. Then $AV=0$, while
\begin{equation}
    \ell_j^\top(E^\star+V)=\lVert v\rVert_2^2u^\top\ne0.
\end{equation}
This gives a solution that violates the law. Hence identifiability is equivalent to $P_0\ell_j=0$, or Equation~\ref{eq:identifiability}.

We measure the unresolved fraction of a nonzero contrast by
\begin{equation}
    \rho_j=\frac{\lVert P_0\ell_j\rVert_2}{\lVert\ell_j\rVert_2}.
    \label{eq:rho}
\end{equation}
Thus $\rho_j=0$ exactly when the law is identifiable. The same argument applies to any linear query contrast, provided a consistent solution gives its intended value. It characterizes the set of solutions, independently of the optimizer. An optimizer may select a particular solution when $\rho_j>0$, and nearest-target retrieval may be correct even with a nonzero geometric residual.

\subsection{Shallow dynamics: proof of Proposition~\ref{prop:dynamics}}
\label{app:shallow}

Differentiating the objective gives
\begin{equation}
    \nabla_E\mathcal{L}=\frac{1}{|D|}A^\top(AE-C)=HE-B.
\end{equation}
To first order in the per-row step size $\eta$, one epoch of SGD sums these row gradients at the epoch's starting state. Its update is $-\eta A^\top(AE-C)$. Taking the continuous-time approximation in epoch units gives $\tau\dot E=B-HE$, with $\tau=(\eta|D|)^{-1}$. Finite-step, order-dependent corrections are omitted by this approximation.

For any consistent $E^\star$, $HE^\star=B$. Projecting the embedding error onto an eigenvector $q_k$ gives the row vector $\boldsymbol\zeta_k(t)=q_k^\top(E(t)-E^\star)\in\mathbb{R}^{1\times d}$, with
\begin{equation}
    \tau\dot{\boldsymbol\zeta}_k=-\lambda_k \boldsymbol\zeta_k,\qquad
    \boldsymbol\zeta_k(t)=\boldsymbol\zeta_k(0)e^{-\lambda_k t/\tau}.
    \label{eq:mode}
\end{equation}
Positive modes decay independently, and zero modes remain at initialization. Although $E^\star$ need not be unique, its positive-mode projections are fixed by the evidence:
\begin{equation}
    q_k^\top E^\star=\lambda_k^{-1}q_k^\top B\quad(\lambda_k>0).
\end{equation}
Thus the positive-mode errors used in Proposition~\ref{prop:dynamics} require no choice of an otherwise unconstrained solution.

For an identifiable law, $\ell_j=P_+\ell_j$ and $\ell_j^\top E^\star=0$. Expanding in the eigenbasis yields
\begin{align}
    \boldsymbol\delta_j(t)
    &=\ell_j^\top(E(t)-E^\star)\nonumber\\
    &=\sum_{k:\lambda_k>0}(\ell_j^\top q_k)\boldsymbol\zeta_k(t)
      =\sum_{k:\lambda_k>0}a_{jk}\boldsymbol\zeta_k(0)e^{-\lambda_k t/\tau},
\end{align}
which proves the proposition. For a non-identifiable law, choosing a consistent $E^\star$ that satisfies the law leaves the additional term $\ell_j^\top P_0(E(0)-E^\star)$, which persists under shallow learning. Several modes can contribute with different signs and rates, so the norm of the law residual need not decay monotonically.

The full continuous-time prediction is
\begin{equation}
    \widehat E(t)=E^\star+P_0(E(0)-E^\star)
      +\sum_{k:\lambda_k>0}q_k \boldsymbol\zeta_k(0)e^{-\lambda_k t/\tau}.
    \label{eq:reconstruction}
\end{equation}
Applying each evaluation query to this matrix retains the evolving source, relation, target, and distractor embeddings. The law residual alone does not specify a nearest-target decision. Predicted event times use the same sampled evaluations and thresholds as the network, without fitting a rate or threshold to its trajectory.

\subsection{Discrete shallow predictions and their relation to SGD}
\label{app:discrete}

The implemented shallow predictor uses the discrete mean update
\begin{equation}
    \widehat E_{t+1}=\widehat E_t-\eta A^\top(A\widehat E_t-C).
    \label{eq:discrete}
\end{equation}
Writing $\sigma_k=|D|\lambda_k$ for the eigenvalues of $A^\top A$, its closed form is
\begin{equation}
    \widehat E_t=E^\star+
      Q\operatorname{diag}\!\bigl((1-\eta\sigma_k)^t\bigr)Q^\top(E(0)-E^\star).
    \label{eq:discrete-solution}
\end{equation}
The code uses the minimum-norm least-squares solution for $E^\star$. All positive-mode errors decay when $0<\eta<2/\sigma_{\max}$; zero-mode components remain unchanged. For small $\eta\sigma_k$, the decay factor approximates $e^{-\eta\sigma_k t}=e^{-\lambda_k t/\tau}$.

Equation~\ref{eq:discrete-solution} is exact for Equation~\ref{eq:discrete}, not for the ordered product of per-row SGD updates. Per-row SGD also preserves the shallow null component exactly, since each update lies in the row space of $A$, but its constrained trajectory can differ at finite step size. The experiments compare these predicted and observed trajectories directly.

\subsection{Factorized depth and unconstrained directions}
\label{app:depth}

For $E=W_N\cdots W_1$, define
\begin{equation}
    L_l=W_N\cdots W_{l+1},\qquad R_l=W_{l-1}\cdots W_1,
\end{equation}
with empty products $L_N=I_{p\times p}$ and $R_1=I_{d\times d}$. A variation in layer $W_l$ gives $dE=L_l\,(dW_l)\,R_l$. The chain rule therefore gives
\begin{equation}
    \nabla_{W_l}\mathcal{L}=L_l^\top(HE-B)R_l^\top.
\end{equation}
Gradient flow yields Equation~\ref{eq:deep}. Because the layer products evolve, the environmental modes generally interact. We integrate the layer equations numerically and reconstruct $\widehat E(t;N)$ at each evaluation point; \appref{app:training} gives the integration settings.

The induced embedding dynamics are
\begin{equation}
    \tau\dot E=\sum_{l=1}^{N}L_lL_l^\top(B-HE)R_l^\top R_l.
    \label{eq:induced}
\end{equation}
Although $P_0(B-HE)=0$, in general $P_0L_lL_l^\top(B-HE)\ne0$. The embedding's null component can consequently evolve. More specifically, $P_0\dot W_N=0$, but $P_0E=(P_0W_N)W_{N-1}\cdots W_1$ can change through the remaining factors. If $P_0W_N(0)=0$, the end-to-end null component instead remains zero.

For example, take $A=(1,0)$, $C=1$, $W_2=(1,1)^\top$, $W_1=2$, and $\tau=1$. Then $B-HE=(-1,0)^\top$, $\dot W_2=(-2,0)^\top$, and $\dot W_1=-1$, giving $\dot E=(-5,-1)^\top$. The unconstrained second entry changes, whereas shallow flow leaves it fixed. This is optimizer-dependent evolution in a direction that the evidence does not identify. Proposition~\ref{prop:identifiability} continues to hold; in our parameterization, factorization does not restrict the set of representable embedding matrices.

\subsection{How a linking fact changes the system}
\label{app:linking}

A new fact contributes one row $g^\top=e_a^\top+e_r^\top-e_b^\top$, with zero target. With no new tokens or anchors, the post-intervention system is
\begin{equation}
    A'=\begin{pmatrix}A\\g^\top\end{pmatrix},\qquad
    C'=\begin{pmatrix}C\\0\end{pmatrix},\qquad
    A'^\top A'=A^\top A+gg^\top.
\end{equation}
Its null space is $\ker A'=\ker A\cap\ker g^\top$. A new row can therefore remove an unconstrained direction only if it has a component along that direction. Its effect is shared by all query contrasts that depend on the removed freedom.

In Experiment~3, the first lattice's anchors and within-lattice facts fix its geometry and the shared offsets. The second lattice's internal facts fix its shape but leave its translation free. If $v$ has entries one on the second lattice's entities and zero elsewhere, this freedom is $E\mapsto E+vu^\top$ for arbitrary $u\in\mathbb{R}^d$, and $\ker A=\operatorname{span}(v)$. The linking row runs from the first lattice to the second, so $g^\top v=-1$ and removes this freedom. Each cross-lattice query has error contrast $c=e_a+n_xe_x+n_ye_y-e_b$, also with $c^\top v=-1$. Hence the link determines all these contrasts, not just its own trained pair. This is a statement about solutions of the constraints; the time needed to retrieve the targets still depends on learning dynamics.

For post-intervention predictions, we recompute $H'=A'^\top A'/(|D|+1)$ and $B'=A'^\top C'/(|D|+1)$. The per-row learning rate is retained, so the epoch time constant becomes $\tau'=(\eta(|D|+1))^{-1}$. Each predicted condition continues from its own predicted pre-intervention state. The network conditions similarly copy their common trained state. Predictions are never reset to the trained network's intervention state.

\clearpage
\section{Experimental Details}
\label{app:methods}

\subsection{Synthetic relational worlds}
\label{app:worlds}

Worlds contained arbitrary entity and relation labels. The primary construction instantiated additive laws $\mathbf{r}_z=\mathbf{r}_x+\mathbf{r}_y$ through facts of the form $a\xrightarrow{x}b$, $b\xrightarrow{y}c$, and $a\xrightarrow{z}c$. Each law appeared across multiple entity sets. Some valid $z$-facts were included in training and distinct valid $z$-facts were withheld as compositional queries. This separates retrieval of an unseen composite fact from memorization of a trained triple.

In the integration experiment, two relational lattices used the same relation types but separate entity sets. Before intervention, all within-lattice constraints could be learned while the lattices' relative translation remained unconstrained. One cross-lattice fact removed that freedom in the linked condition. The matched no-link condition continued from the identical network state without receiving the resolving constraint.

\paragraph{Experiment 1: law-specific emergence.}
The world uses eight lattice specifications $(m,n,r_x,r_y)$, each instantiated for two laws with distinct relation tokens: $(5,4,1,1)$, $(4,4,2,2)$, $(4,4,3,4)$, $(4,4,4,6)$, $(5,4,2,8)$, $(4,4,1,4)$, $(4,5,4,8)$, and $(4,4,3,6)$. Here $m,n$ are the numbers of entity positions along the two axes, indexed by $0\le i<m$ and $0\le j<n$. The values $r_x,r_y$ count presentations of each corresponding edge per epoch; trained diagonal edges use the same multiplicity as the second-axis edges. The $(m-1)(n-1)$ diagonal cells correspond to composite facts $(i,j)\xrightarrow{z}(i+1,j+1)$. The constructor samples 25\% without replacement for training, rounded to an integer and with at least one included. Each lattice has unit-weight anchors at $(0,0)$, $(m-1,0)$, and $(0,n-1)$, using its generated target coordinates: 48 anchors across the 16 lattices. The world specifications were selected during exploratory development to separate emergence times; they are not a random sample of relational topologies. The three displayed laws are \texttt{L3\_1}, \texttt{L2\_1}, and \texttt{L5\_0}, selected once from pooled results and then held fixed; all 16 laws enter the reported statistics.

\paragraph{Experiment 2: structural identifiability.}
Each world has two blocks, each with 16 premise triads and four composite facts. In the closed block, the four composite facts connect the endpoints of the first four observed premise triads. In the open block, they instead connect four additional entity pairs sharing no entities with any premise triad; the premise triads themselves remain connected two-edge paths. Each block anchors the initial entity of every premise triad and the middle and final entities of its first triad (18 anchors per block). The open block also anchors the head of each additional pair, leaving its tail unanchored: 40 unit-weight anchors in total, all at generated target coordinates. The intervention adds the composite fact between the endpoints of the first open-block premise triad; no anchor is added. Both assignments of open and closed blocks are run for every world. The evaluation set excludes facts present after the intervention and is held fixed across conditions.

\paragraph{Experiment 3: relational integration.}
Each world uses two $3\times3$ lattices sharing offsets $x,y,z=x+y$, with three anchor targets in the first lattice and none in the second. The added fact connects the lattices using the existing $x$ relation. The offsets and initial lattice origins are Gaussian draws in 16 dimensions. The second lattice is then translated so that its origin lies one $x$ step beyond the first lattice's final first-axis entity; this makes the linking fact true in the constructed world. The first lattice's anchors are at $(0,0)$, $(1,0)$, and $(0,1)$.

\subsection{Training and numerical prediction}
\label{app:training}

The model minimized Equation~\ref{eq:loss} using SGD with a random permutation of all fact and anchor rows in every epoch, a fixed per-row learning rate, and no adaptive optimizer. Shallow networks learned $E$ directly. Deeper networks used $E=W_N\cdots W_1$, with $W_N\in\mathbb{R}^{p\times d}$ and the other layers in $\mathbb{R}^{d\times d}$. Here $d=16$, so the factorization does not restrict the class of embedding matrices. All initial entries were Gaussian with zero mean and standard deviation 0.01 for shallow embeddings and 0.05 for each deep layer.

The selected worlds have indices $s=0,\ldots,199$, with model initialization seed $1000+s$. World generation uses seed $s$ in Experiments~1 and 3 and $1000+s$ in Experiment~2. Thus the Experiment~2 world generator and model initializer use the same seed in separate generator instances; their random streams are not independent within a world. Experiment~1 uses presentation-order seed $2000+s$. Experiments~2 and 3 use order seed 7 before intervention and 8 after intervention, shared across worlds; their uncertainty summaries therefore vary world draws and initializations, not order seeds independently.

Shallow predictions use Equation~\ref{eq:discrete-solution}; deeper predictions use fourth-order Runge--Kutta integration of Equation~\ref{eq:deep}, with the depth-specific substeps in \papertabref{tab:configuration}. Predictions start from the same initial weights as training and use the same fixed query set. At intervention, the two network conditions copy the network's pre-intervention state, while the two predicted conditions continue from their own predicted pre-intervention state. The resolving fact is one additional distinct training constraint, revisited each subsequent epoch alongside the original constraints; this is not a single-update experiment. Pretraining durations were chosen separately by depth using settling diagnostics, and post-intervention trajectories are indexed by epochs since the added fact.

\input{experimental_configuration.tex}

\subsection{Held-out queries and evaluation measures}
\label{app:evaluation}

In the emergence and identifiability experiments, each held-out query was an unseen composite fact $a\xrightarrow{z}b$, evaluated at
\begin{equation}
    \mathbf{q}=\mathbf{x}_a+\mathbf{r}_z.
\end{equation}
Candidates were restricted to entities in the corresponding lattice (Experiment~1) or block (Experiment~2). Compositional accuracy was the proportion of targets ranked first by Euclidean distance. Geometric error was computed with Equation~\ref{eq:geo}.

In the integration experiment, held-out items were cross-structure comparisons. If source $a$ and target $b$ were separated by $n_x$ steps along relation $x$ and $n_y$ steps along relation $y$, the query was
\begin{equation}
    \mathbf{q}=\mathbf{x}_a+n_x\mathbf{r}_x+n_y\mathbf{r}_y.
\end{equation}

The integers $n_x,n_y$ are supplied query coefficients derived from the constructed world's coordinates. The task tests their application to learned offsets, rather than discovery of the path. Retrieval uses all 18 entities across the two lattices. Of the 81 cross-lattice pairs, the directly trained linking pair is excluded, leaving 80 queries. Geometric error uses Equation~\ref{eq:geo}, with spacing computed among the nine destination entities. Thus the retrieval pool includes distractors from the source lattice while the geometric reference scale remains that of the destination lattice.

For $M$ held-out queries, geometric error is
\begin{equation}
    \mathcal{E}_{\mathrm{geo}}=
    \frac{1}{M}\sum_{i=1}^{M}
    \frac{\lVert\mathbf{q}_i-\mathbf{x}_{b_i}\rVert_2}{s_i+10^{-12}},
    \label{eq:geo}
\end{equation}
where $s_i$ is the median nearest-neighbour distance among the learned entity embeddings in query $i$'s reference structure, recomputed at each evaluation. The reference structure is the retrieval candidate set in Experiments~1 and 2 and the nine-entity destination lattice in Experiment~3. The denominator expresses error in units of the target structure's current spacing; $10^{-12}$ prevents division by zero. Retrieval is counted as correct if no candidate is strictly closer than the target, matching the implementation's treatment of ties.

Full-pool performance before linking was low on average but not zero in every world. At intervention, 177, 198, and 198 of 200 worlds answered none of the 80 queries correctly at depths 1--3. The largest initial scores were 9, 27, and 63, respectively. Without a link, depth-1 world 1 retained its score of 9 through 30,000 epochs; the depth-2 and depth-3 maxima both occurred in world 91 and fell to zero by that time. We report raw counts and matched no-link controls rather than subtracting a baseline.

\subsection{Emergence criteria and summary statistics}
\label{app:statistics}

For Experiment~1, $t^\ast$ was the first evaluation at which held-out accuracy equalled 100\% and normalized geometric error was below 0.5, with both criteria sustained for ten consecutive evaluations. Experiment~2 recovery was the first evaluation reaching 90\% held-out accuracy. Experiment~3 unlock was the first evaluation with at least 72 of 80 comparisons correct among all 18 candidates. The latter two criteria do not require a subsequent period of sustained success.

Timing agreement is $R^2=1-\sum_i(y_i-\widehat y_i)^2/\sum_i(y_i-\bar y)^2$, where $y_i=\log_{10}t_i$ and only pairs of finite, positive event times are included. This measures agreement without fitting a regression; it is not squared correlation. Events attained at intervention contribute to attainment counts and absolute errors but have no logarithmic time. This excludes one depth-1 world from Experiment~2's log-time comparison. All Experiment~3 event times are positive.

Experiments~1 and 2 use the original evaluation intervals in \papertabref{tab:configuration}. Experiment~3 uses the denser replay grid described in \appref{app:controls}, with predictions evaluated at every network evaluation time. The observed crossing interval is the gap from the last check below criterion to the first meeting it. The reported within-interval share counts worlds in which the absolute difference between observed and predicted first-success times is no greater than that world's observed crossing interval. This is a tolerance on sampled event times, not a test of exact continuous-time coincidence.

At depths 1--3, median Experiment~3 crossing intervals are 943, 155, and 58.5 epochs, with maxima 1,779, 424, and 250. Median absolute timing error is zero at each depth; the 95th percentiles are 0, 144.8, and 202.5 epochs. The within-interval shares are 99.5\%, 100.0\%, and 95.0\%. These finite-resolution measures accompany the log-time $R^2$ in Section~\ref{sec:result3}.

Trajectory $R^2$ uses raw accuracy or $\log_{10}$ geometric error, computed within each world (and law in Experiment~1) before averaging; constant trajectories with undefined $R^2$ are omitted. Each evaluation carries equal weight. Experiment~3 trajectory agreement uses the shared replay grid over the first 30,000 post-intervention epochs; its denser early sampling gives that part of the trajectory more weight. Experiment~1 uses the full depth-specific runs, and Experiment~2 uses the first 2,000 post-intervention epochs for trajectory agreement and effects. Experiment~3 effects are reported at 30,000 epochs; unlock times use the full 40,000-epoch follow-up. Worlds are the replication unit, with the two Experiment~2 structural assignments averaged within each world before inference. Reported confidence intervals are 95\% Student-$t$ intervals over the 200 world-level values; fold-change intervals are computed in log space and exponentiated. Accuracy means are arithmetic; geometric curves use geometric means and log-space standard errors.

\input{pat_control_note.tex}

\clearpage
\section{Supplementary Results}
\label{app:results}

This appendix collects full-run emergence times (\appref{app:timing}), premise retention (\appref{app:retention}), geometric trajectories (\appref{app:geometry}), and prediction agreement (\appref{app:agreement}). Figures and tables are numbered by appendix, so each can be cited directly.

\subsection{Full-horizon emergence times}
\label{app:timing}

\begin{figure}[htbp]
    \centering
    \includegraphics[width=\linewidth]{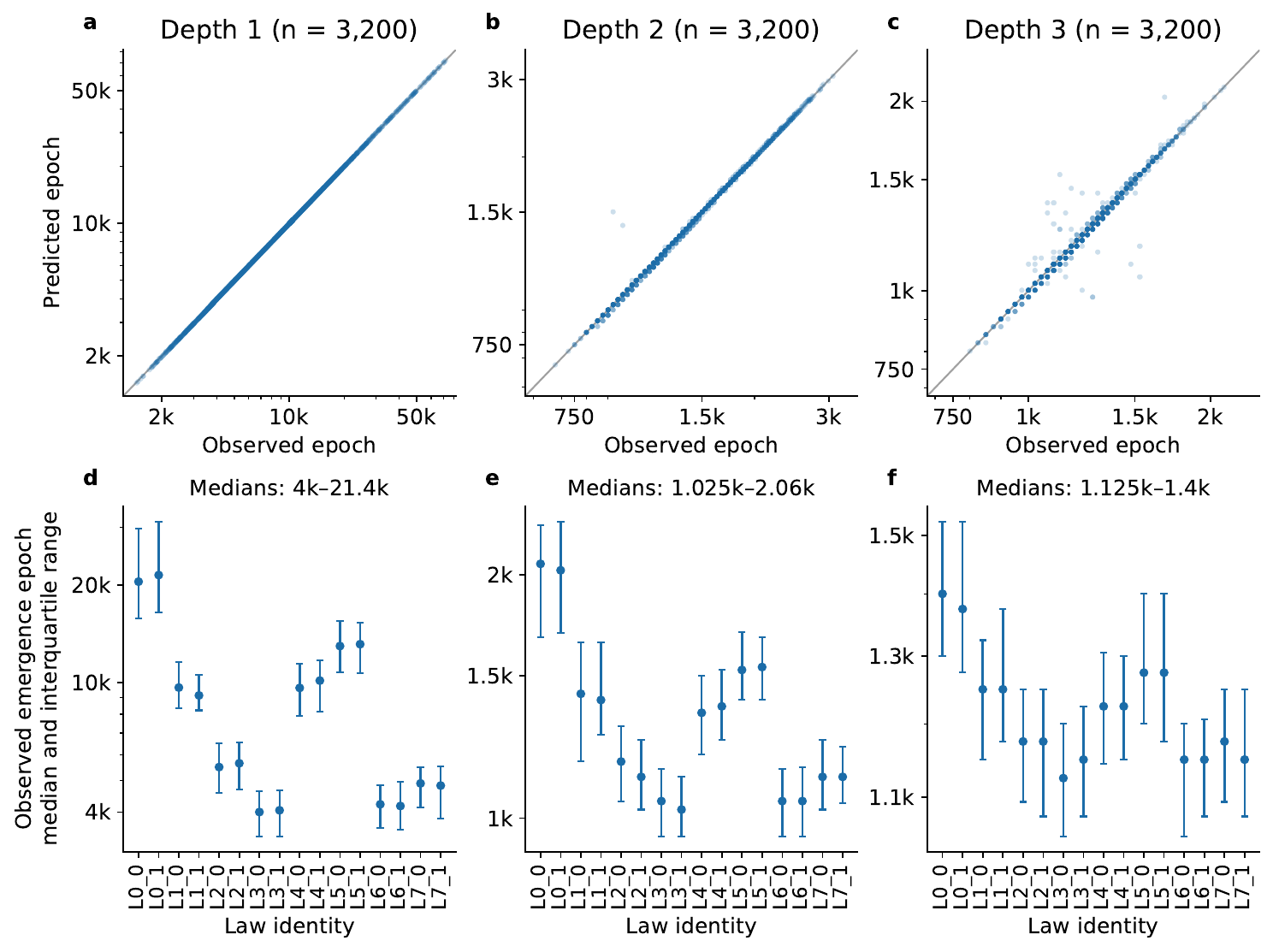}
    \caption{Emergence times across all 16 laws and 200 worlds. \textbf{a--c}, predicted versus observed emergence epochs for all 3,200 law-worlds at each depth; the diagonal denotes equality. Points overlap when event times coincide on the evaluation grid. \textbf{d--f}, median observed event time for each fixed law, with interquartile ranges across worlds. All timing axes are logarithmic; limits differ across depths. Error bars describe between-world variation, not uncertainty in the median.}
    \label{fig:timing}
\end{figure}

\paperfigref{fig:timing} shows every law-world's event time and the distribution of emergence times across worlds for each law. Unlike the early trajectories in \paperfigref{fig:emergence}, these summaries include the full 80,000-epoch shallow run and 10,000-epoch deep runs. All 3,200 law-worlds reach criterion at each depth, so none are omitted for censoring. A fixed law label fixes its lattice dimensions and presentation counts, not its sampled diagonal constraints, target vectors, or initial projections. These draws alter the environmental spectrum, multimode coefficients, and retrieval geometry. The per-world predictor incorporates them together; the present experiment does not isolate their individual contributions to timing variance.

\clearpage
\subsection{Premise retention during integration}
\label{app:retention}

\begin{figure}[htbp]
    \centering
    \includegraphics[width=\linewidth]{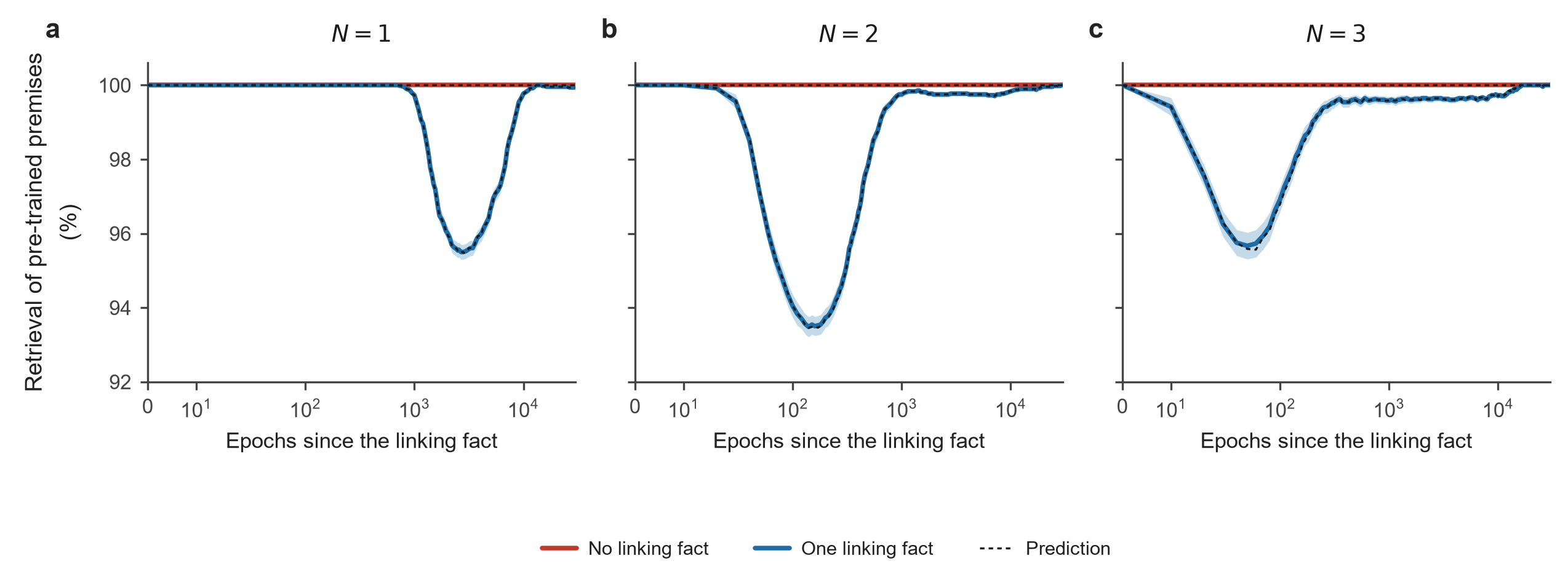}
    \caption{Linking two structures can temporarily disrupt retrieval of original premises. Mean retrieval of the 24 original $x/y$ premise facts in Experiment~3, scored within each fact's own lattice, over the same 200 worlds per depth as \paperfigref{fig:integration}. The ordinate refers to these premises, not all 32 original training facts. Blue and red curves show linked and no-link networks; dashed curves show theory predictions without trajectory-specific fitting. Shading is s.e.m. across worlds. Time is linear below ten epochs and logarithmic above. The first 30,000 post-intervention epochs are shown; recovery assessments use the full 40,000-epoch run. Curves connect sampled evaluations. Mean retention can approach 100\% while an individual world still has an error.}
    \label{fig:retention}
\end{figure}

\paperfigref{fig:retention} shows the transient premise-retrieval losses described in Section~\ref{sec:result3}. The sample, candidate sets, and recovery criteria are specified in \appref{app:controls}.

\clearpage
\subsection{Geometric trajectories}
\label{app:geometry}

\begin{figure}[htbp]
    \centering
    \includegraphics[width=\linewidth]{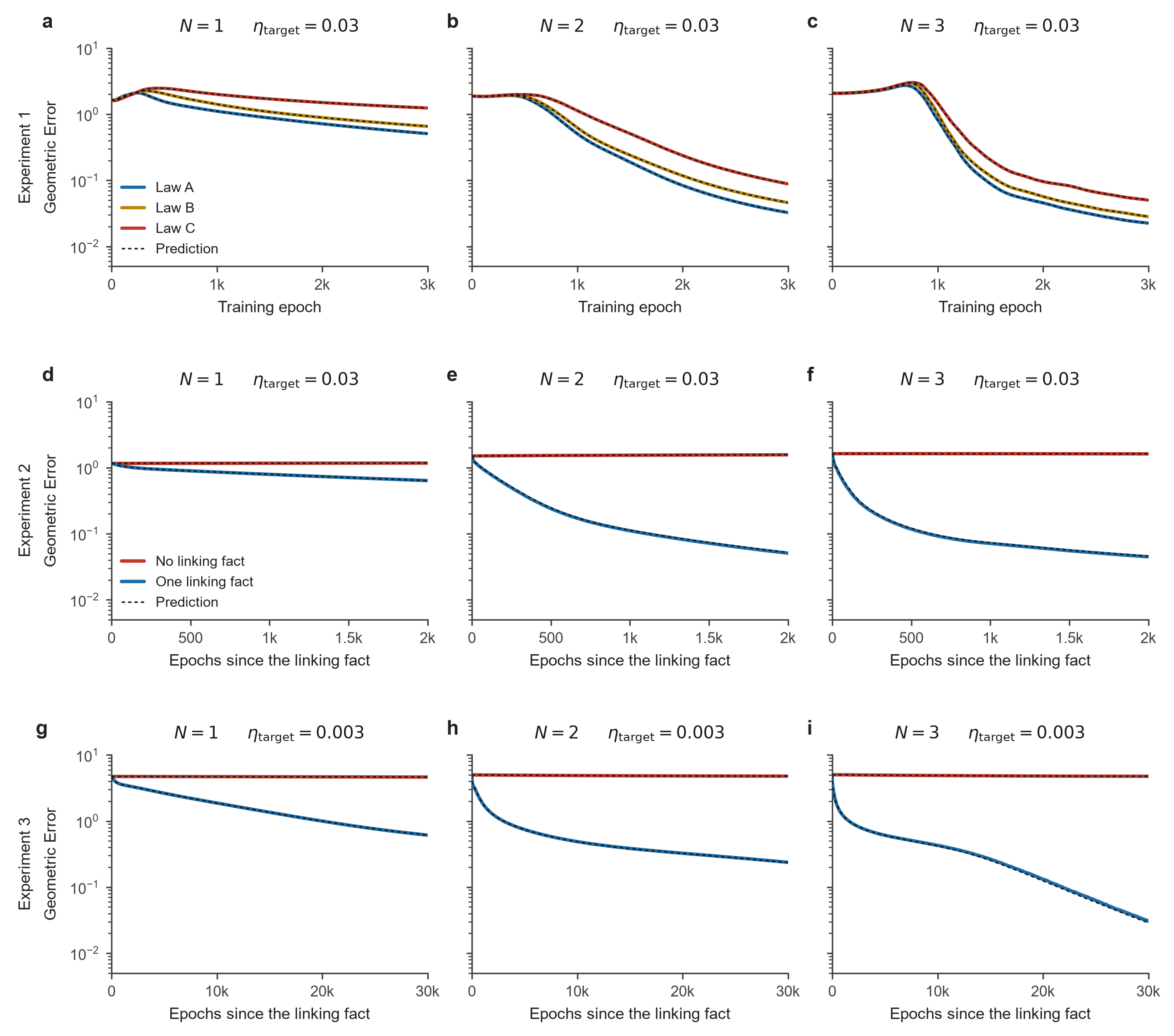}
    \caption{Geometric error in the three experiments. \textbf{a--c}, Experiment~1, and \textbf{d--f}, Experiment~2: query-to-target distance normalized by median nearest-neighbour spacing among candidates. \textbf{g--i}, Experiment~3: distance between each cross-structure query and its correct target, normalized by median nearest-neighbour spacing among the nine destination entities. Curves are geometric means across 200 worlds, shaded by log-space s.e.m.; dashed lines show predictions.}
    \label{fig:geometry}
\end{figure}

\paperfigref{fig:geometry} shows the geometric error behind Figures~\ref{fig:emergence}--\ref{fig:integration}, one row per experiment. Each row uses the same worlds, runs and predictions as the corresponding main-text figure, with the geometric error defined in Section~\ref{sec:measures}.

\clearpage
\subsection{Prediction agreement}
\label{app:agreement}

\papertabref{tab:trajectories} compares predicted and observed trajectories over each experiment's full horizon, using the per-trajectory $R^2$ defined in \appref{app:statistics}.

\begin{table}[htbp]
\caption{Agreement between predicted and observed trajectories. Values are means over worlds (over law-worlds in Experiment~1) of the per-trajectory $R^2$, and for Experiment~2 also medians over worlds of the mean absolute accuracy difference and of the geometric-error ratio. Experiments~2 and 3 report the linked condition; no-link trajectories are nearly flat, and at the 95th percentile their predictions differ by at most 0.17 percentage points in accuracy and a factor of 1.002 in geometric error. Horizons are 80,000 epochs at $N=1$ and 10,000 at $N=2,3$ in Experiment~1, and 2,000 and 30,000 epochs after the linking fact in Experiments~2 and 3.}
\label{tab:trajectories}
\centering
\begin{tabular}{llccc}
\toprule
Experiment & Measure & $N=1$ & $N=2$ & $N=3$ \\
\midrule
1 & Accuracy, mean $R^2$ & 0.9999 & 0.9997 & 0.9989 \\
  & Log geometric error, mean $R^2$ & 1.0000 & 1.0000 & 0.9999 \\
\addlinespace
2 & Accuracy, mean $R^2$ & 0.9978 & 0.9808 & 0.9359 \\
  & Accuracy, median absolute error (pp) & 0.00 & 0.22 & 0.25 \\
  & Log geometric error, mean $R^2$ & 1.0000 & 0.9994 & 0.9977 \\
  & Geometric error, median ratio & 1.000 & 1.020 & 1.027 \\
\addlinespace
3 & Correct comparisons, mean $R^2$ & 1.0000 & 0.9999 & 0.9992 \\
  & Log geometric error, mean $R^2$ & 1.0000 & 1.0000 & 0.9993 \\
\bottomrule
\end{tabular}
\end{table}

Two thousand epochs after the intervention in Experiment~2, linked accuracy exceeded no-link accuracy by 66.5, 96.8, and 97.4 percentage points at $N=1,2,3$ (95\% CIs 61.7--71.3, 95.3--98.3, and 96.0--98.7). Thirty thousand epochs after the link in Experiment~3, linked networks answered 77.8, 79.9, and 80.0 of 80 comparisons at $N=1,2,3$ (95\% CIs 77.2--78.5, 79.8--80.0, and 80.0--80.0), with all 80 correct in 129, 196, and 200 of 200 worlds, and geometric error fell 7.73-, 21.0-, and 163-fold (95\% CIs 7.46--8.02, 19.9--22.2, and 137--194).

In Experiment~2, the recovery-time $R^2$ is lower at $N=3$ (0.8735) than at $N=2$ (0.9860). This is not solely a low-variance artifact: from $N=2$ to $N=3$ the observed log-time variance falls from 0.0311 to 0.0267, while the mean squared log-time error rises from 0.000435 to 0.00337. Absolute agreement on the evaluation grid and relative timing precision therefore give complementary assessments. At $N=1$, one world already above 90\% accuracy at the intervention is excluded from $R^2$.

\end{document}

%% file: math_commands.tex
\usepackage{amsmath,amsfonts,bm}

\def\eqref#1{equation~\ref{#1}}

\def\1{\bm{1}}

\DeclareMathAlphabet{\mathsfit}{\encodingdefault}{\sfdefault}{m}{sl}
\SetMathAlphabet{\mathsfit}{bold}{\encodingdefault}{\sfdefault}{bx}{n}



%% file: experimental_configuration.tex
\begin{table}[t]
\centering
\caption{Run settings for the selected 200 worlds per depth. One epoch visits every constraint row once, including repeated facts and anchors. For staged runs, $\eta=\eta_{\mathrm{target}}/\lambda_{\max}(A_0^\top A_0)$ is set using the pre-intervention matrix $A_0$ and retained after the intervention. Shallow predictions use the discrete closed form. The original intervals are supplemented by denser replay evaluations for Experiment~3 and the retention controls (\appref{app:controls}).}
\label{tab:configuration}
\small
\begin{tabular}{llrrr}
\toprule
Experiment & Setting & $N=1$ & $N=2$ & $N=3$ \\
\midrule
1 & $\eta_{\mathrm{target}}$ & 0.03 & 0.03 & 0.03 \\
  & Total epochs & 80,000 & 10,000 & 10,000 \\
  & Original evaluation interval & 25 & 25 & 25 \\
  & RK4 substeps per epoch & --- & 4 & 8 \\
\midrule
2 & $\eta_{\mathrm{target}}$ & 0.03 & 0.03 & 0.03 \\
  & Pre-intervention epochs & 50,000 & 6,000 & 3,500 \\
  & Post-intervention epochs & 40,000 & 8,000 & 4,000 \\
  & Original evaluation interval & 50 & 10 & 10 \\
  & RK4 substeps per epoch & --- & 4 & 8 \\
\midrule
3 & $\eta_{\mathrm{target}}$ & 0.003 & 0.003 & 0.003 \\
  & Pre-intervention epochs & 200,000 & 65,000 & 55,000 \\
  & Post-intervention epochs & 40,000 & 40,000 & 40,000 \\
  & Original evaluation interval & 2,500 & 500 & 250 \\
  & RK4 substeps per epoch & --- & 1 & 1 \\
\bottomrule
\end{tabular}
\end{table}

The normalized learning rate is $\eta_{\mathrm{target}}=\eta\lambda_{\max}(A^\top A)$. Positive modes are selected using an eigenvalue cutoff of $10^{-9}$ times the largest eigenvalue; contrasts with $\rho<10^{-7}$ are classified as identifiable. The observed ambiguity values are separated by many orders of magnitude from this cutoff. The candidate-spacing denominator includes $10^{-12}$ to avoid division by zero.

%% file: pat_control_note.tex
\subsection{Replay validation and retention controls}
\label{app:controls}

We replayed all 200 Experiment~3 worlds at each depth from their original seeds, preserving initialization, presentation order, learning rates, and durations. At every original post-intervention evaluation, the replay reproduced the stored held-out hits and geometric scores exactly. It additionally scored all 18 retrieval candidates and the original training facts. Theory predictions reproduced the stored predicted hits exactly and geometric scores to relative tolerance $10^{-9}$.

The replay grid combines the original evaluation times with checks every ten epochs through epoch 500, every 50 through epoch 2,000, and a geometrically spaced sequence thereafter with factor 1.05, rounded to integer epochs. Over 40,000 post-intervention epochs this gives 158, 218, and 294 evaluations at depths 1--3. Predictions include every network evaluation time; finer prediction-only points are used to draw the curves but not to compare event times. Pre-intervention training lasts 200,000, 65,000, and 55,000 epochs, respectively.

Retention tests retrieval of an original fact using candidates in its own structure. In Experiment~3 we score both the 24 original $x/y$ premises and all 32 original training facts, excluding the newly added link. All are correctly retrieved at intervention. A distinct loss counts a fact that fails at any sampled evaluation; a simultaneous loss counts facts failing at the same evaluation. These differ when different facts fail at different times.

At least one premise fails in 162, 170, and 136 linked worlds at depths 1--3, and in no no-link world. Among affected worlds, the median numbers of distinct premise losses are 1, 2, and 2, with maxima 4, 10, and 14. Maximum simultaneous premise losses are 3, 10, and 10. Scoring all 32 original facts gives the same affected-world counts, but distinct-loss medians of 1, 3, and 2 and maxima of 5, 11, and 16. The maximum simultaneous losses across all original facts are 4, 11, and 11. Of 468 affected world--depth cases, 467 return to full retrieval at all subsequent evaluations through 40,000 epochs; depth-1 world 22 has not recovered by that horizon. These observations do not exclude losses between checks or later failures.

The theory predicts premise losses in 162, 171, and 135 worlds. Its affected set matches the network at depth 1, adds world 144 at depth 2, and misses world 10 at depth 3. The minima of the network's mean premise-retention curves are 95.50\%, 93.48\%, and 95.67\%; the corresponding theoretical minima on the shared grids are 95.48\%, 93.46\%, and 95.56\%. These minima describe retention, not prediction accuracy. The maximum pointwise gaps between the mean curves are 0.021, 0.104, and 0.167 percentage points.

For Experiment~2, we replayed seeds 0--29 at each depth with both assignments of the open and closed blocks. Each of the 180 cells included linked and no-link continuations. All 64 original premises and all 72 original training facts remained correct at every evaluation in both conditions. The depth-specific post-intervention horizons were 40,000, 8,000, and 4,000 epochs, sampled at 899, 824, and 412 evaluations. Every original evaluation within those horizons was reproduced exactly. The result is no detected loss in this sample and on these grids, rather than a guarantee of stability between evaluations.

%% file: main.bbl
\begin{thebibliography}{24}
\providecommand{\natexlab}[1]{#1}
\providecommand{\url}[1]{\texttt{#1}}
\expandafter\ifx\csname urlstyle\endcsname\relax
  \providecommand{\doi}[1]{doi: #1}\else
  \providecommand{\doi}{doi: \begingroup \urlstyle{rm}\Url}\fi

\bibitem[Behrens et~al.(2018)Behrens, Muller, Whittington, Mark, Baram,
  Stachenfeld, and Kurth-Nelson]{behrens2018cognitive}
Timothy E.~J. Behrens, Timothy~H. Muller, James C.~R. Whittington, Shirley
  Mark, Alon~B. Baram, Kimberly~L. Stachenfeld, and Zeb Kurth-Nelson.
\newblock What is a cognitive map? organizing knowledge for flexible behavior.
\newblock \emph{Neuron}, 100\penalty0 (2):\penalty0 490--509, 2018.
\newblock \doi{10.1016/j.neuron.2018.10.002}.

\bibitem[Bordes et~al.(2013)Bordes, Usunier, Garcia-Duran, Weston, and
  Yakhnenko]{bordes2013translating}
Antoine Bordes, Nicolas Usunier, Alberto Garcia-Duran, Jason Weston, and Oksana
  Yakhnenko.
\newblock Translating embeddings for modeling multi-relational data.
\newblock In \emph{Advances in Neural Information Processing Systems},
  volume~26, 2013.

\bibitem[Cohen et~al.(2024)Cohen, Biran, Yoran, Globerson, and
  Geva]{cohen2024evaluating}
Roi Cohen, Eden Biran, Ori Yoran, Amir Globerson, and Mor Geva.
\newblock Evaluating the ripple effects of knowledge editing in language
  models.
\newblock \emph{Transactions of the Association for Computational Linguistics},
  12:\penalty0 283--298, 2024.
\newblock \doi{10.1162/tacl_a_00644}.
\newblock URL \url{https://aclanthology.org/2024.tacl-1.16/}.

\bibitem[Dekker et~al.(2022)Dekker, Otto, and
  Summerfield]{dekker2022curriculum}
Ronald~B. Dekker, Fabian Otto, and Christopher Summerfield.
\newblock Curriculum learning for human compositional generalization.
\newblock \emph{Proceedings of the National Academy of Sciences}, 119\penalty0
  (41):\penalty0 e2205582119, 2022.
\newblock \doi{10.1073/pnas.2205582119}.

\bibitem[Dettmers et~al.(2018)Dettmers, Minervini, Stenetorp, and
  Riedel]{dettmers2018convolutional}
Tim Dettmers, Pasquale Minervini, Pontus Stenetorp, and Sebastian Riedel.
\newblock Convolutional {2D} knowledge graph embeddings.
\newblock In \emph{AAAI Conference on Artificial Intelligence}, 2018.
\newblock URL \url{https://arxiv.org/abs/1707.01476}.

\bibitem[Frankland \& Greene(2020)Frankland and Greene]{frankland2020two}
Steven~M. Frankland and Joshua~D. Greene.
\newblock Two ways to build a thought: Distinct forms of compositional semantic
  representation across brain regions.
\newblock \emph{Cerebral Cortex}, 30\penalty0 (6):\penalty0 3838--3855, 2020.
\newblock \doi{10.1093/cercor/bhaa001}.

\bibitem[Guu et~al.(2015)Guu, Miller, and Liang]{guu2015traversing}
Kelvin Guu, John Miller, and Percy Liang.
\newblock Traversing knowledge graphs in vector space.
\newblock In \emph{Proceedings of the 2015 Conference on Empirical Methods in
  Natural Language Processing}, pp.\  318--327, 2015.
\newblock \doi{10.18653/v1/D15-1038}.
\newblock URL \url{https://aclanthology.org/D15-1038/}.

\bibitem[Halford et~al.(2010)Halford, Wilson, and
  Phillips]{halford2010relational}
Graeme~S. Halford, William~H. Wilson, and Steven Phillips.
\newblock Relational knowledge: The foundation of higher cognition.
\newblock \emph{Trends in Cognitive Sciences}, 14\penalty0 (11):\penalty0
  497--505, 2010.
\newblock \doi{10.1016/j.tics.2010.08.005}.

\bibitem[Kumaran \& McClelland(2012)Kumaran and
  McClelland]{kumaran2012generalization}
Dharshan Kumaran and James~L. McClelland.
\newblock Generalization through the recurrent interaction of episodic
  memories: A model of the hippocampal system.
\newblock \emph{Psychological Review}, 119\penalty0 (3):\penalty0 573--616,
  2012.
\newblock \doi{10.1037/a0028681}.

\bibitem[Lampinen \& Ganguli(2019)Lampinen and Ganguli]{lampinen2019analytic}
Andrew~K. Lampinen and Surya Ganguli.
\newblock An analytic theory of generalization dynamics and transfer learning
  in deep linear networks.
\newblock In \emph{International Conference on Learning Representations}, 2019.
\newblock URL \url{https://arxiv.org/abs/1809.10374}.

\bibitem[Liu et~al.(2022)Liu, Kitouni, Nolte, Michaud, Tegmark, and
  Williams]{liu2022understanding}
Ziming Liu, Ouail Kitouni, Niklas Nolte, Eric~J. Michaud, Max Tegmark, and Mike
  Williams.
\newblock Towards understanding grokking: An effective theory of representation
  learning.
\newblock In \emph{Advances in Neural Information Processing Systems},
  volume~35, 2022.

\bibitem[Nanda et~al.(2023)Nanda, Chan, Lieberum, Smith, and
  Steinhardt]{nanda2023progress}
Neel Nanda, Lawrence Chan, Tom Lieberum, Jess Smith, and Jacob Steinhardt.
\newblock Progress measures for grokking via mechanistic interpretability.
\newblock In \emph{International Conference on Learning Representations}, 2023.
\newblock URL \url{https://arxiv.org/abs/2301.05217}.

\bibitem[Nelli et~al.(2023)Nelli, Braun, Dumbalska, Saxe, and
  Summerfield]{nelli2023neural}
Stephanie Nelli, Lukas Braun, Tsvetomira Dumbalska, Andrew Saxe, and
  Christopher Summerfield.
\newblock Neural knowledge assembly in humans and neural networks.
\newblock \emph{Neuron}, 111\penalty0 (9):\penalty0 1504--1516.e9, 2023.
\newblock \doi{10.1016/j.neuron.2023.02.014}.

\bibitem[Park et~al.(2024)Park, Okawa, Lee, Tanaka, and
  Lubana]{park2024emergence}
Core~Francisco Park, Maya Okawa, Andrew Lee, Hidenori Tanaka, and Ekdeep~Singh
  Lubana.
\newblock Emergence of hidden capabilities: Exploring learning dynamics in
  concept space.
\newblock In \emph{Advances in Neural Information Processing Systems},
  volume~37, 2024.
\newblock URL \url{https://arxiv.org/abs/2406.19370}.

\bibitem[Saxe et~al.(2014)Saxe, McClelland, and Ganguli]{saxe2014exact}
Andrew~M. Saxe, James~L. McClelland, and Surya Ganguli.
\newblock Exact solutions to the nonlinear dynamics of learning in deep linear
  neural networks.
\newblock In \emph{International Conference on Learning Representations}, 2014.
\newblock URL \url{https://arxiv.org/abs/1312.6120}.

\bibitem[Saxe et~al.(2019)Saxe, McClelland, and Ganguli]{saxe2019mathematical}
Andrew~M. Saxe, James~L. McClelland, and Surya Ganguli.
\newblock A mathematical theory of semantic development in deep neural
  networks.
\newblock \emph{Proceedings of the National Academy of Sciences}, 116\penalty0
  (23):\penalty0 11537--11546, 2019.
\newblock \doi{10.1073/pnas.1820226116}.

\bibitem[Schug et~al.(2024)Schug, Kobayashi, Akram, Wo{\l}czyk, Proca, von
  Oswald, Pascanu, Sacramento, and Steger]{schug2024discovering}
Simon Schug, Seijin Kobayashi, Yassir Akram, Maciej Wo{\l}czyk, Alexandra
  Proca, Johannes von Oswald, Razvan Pascanu, Jo{\~a}o Sacramento, and Angelika
  Steger.
\newblock Discovering modular solutions that generalize compositionally.
\newblock In \emph{International Conference on Learning Representations}, 2024.

\bibitem[Searle(1971)]{searle1971linear}
Shayle~R. Searle.
\newblock \emph{Linear Models}.
\newblock John Wiley \& Sons, New York, 1971.

\bibitem[Sun et~al.(2019)Sun, Deng, Nie, and Tang]{sun2019rotate}
Zhiqing Sun, Zhi-Hong Deng, Jian-Yun Nie, and Jian Tang.
\newblock {RotatE}: Knowledge graph embedding by relational rotation in complex
  space.
\newblock In \emph{International Conference on Learning Representations}, 2019.
\newblock URL \url{https://arxiv.org/abs/1902.10197}.

\bibitem[Toutanova \& Chen(2015)Toutanova and Chen]{toutanova2015observed}
Kristina Toutanova and Danqi Chen.
\newblock Observed versus latent features for knowledge base and text
  inference.
\newblock In \emph{Proceedings of the 3rd Workshop on Continuous Vector Space
  Models and their Compositionality}, pp.\  57--66, 2015.
\newblock \doi{10.18653/v1/W15-4007}.
\newblock URL \url{https://aclanthology.org/W15-4007/}.

\bibitem[Wiedemer et~al.(2023)Wiedemer, Mayilvahanan, Bethge, and
  Brendel]{wiedemer2023compositional}
Thadd{\"a}us Wiedemer, Prasanna Mayilvahanan, Matthias Bethge, and Wieland
  Brendel.
\newblock Compositional generalization from first principles.
\newblock In \emph{Advances in Neural Information Processing Systems},
  volume~36, 2023.
\newblock URL \url{https://arxiv.org/abs/2307.05596}.

\bibitem[Yang et~al.(2025)Yang, Park, Lubana, Okawa, Hu, and
  Tanaka]{yang2025swing}
Yongyi Yang, Core~Francisco Park, Ekdeep~Singh Lubana, Maya Okawa, Wei Hu, and
  Hidenori Tanaka.
\newblock Swing-by dynamics in concept learning and compositional
  generalization.
\newblock In \emph{International Conference on Learning Representations}, 2025.
\newblock URL \url{https://arxiv.org/abs/2410.08309}.

\bibitem[Zeithamova \& Preston(2010)Zeithamova and
  Preston]{zeithamova2010flexible}
Dagmar Zeithamova and Alison~R. Preston.
\newblock Flexible memories: Differential roles for medial temporal lobe and
  prefrontal cortex in cross-episode binding.
\newblock \emph{The Journal of Neuroscience}, 30\penalty0 (44):\penalty0
  14676--14684, 2010.
\newblock \doi{10.1523/JNEUROSCI.3250-10.2010}.

\bibitem[Zhong et~al.(2023)Zhong, Wu, Manning, Potts, and
  Chen]{zhong2023mquake}
Zexuan Zhong, Zhengxuan Wu, Christopher Manning, Christopher Potts, and Danqi
  Chen.
\newblock {MQuAKE}: Assessing knowledge editing in language models via
  multi-hop questions.
\newblock In \emph{Proceedings of the 2023 Conference on Empirical Methods in
  Natural Language Processing}, pp.\  15686--15702, 2023.
\newblock \doi{10.18653/v1/2023.emnlp-main.971}.
\newblock URL \url{https://aclanthology.org/2023.emnlp-main.971/}.

\end{thebibliography}
